\pdfoutput=1
\documentclass[twocolumn,letterpaper,10pt]{article}
\usepackage[textwidth=6.875in,textheight=9.25in,columnsep=0.25in,centering]{geometry}
\usepackage{newtxtext,newtxmath}
\usepackage[hyphens]{url}
\usepackage{graphicx}
\usepackage{natbib}
\usepackage{caption}
\usepackage{booktabs}
\usepackage{tikz}
\usetikzlibrary{arrows.meta,positioning,calc,shapes.arrows}
\definecolor{okgreen}{RGB}{27,120,55}
\definecolor{warnorange}{RGB}{200,105,20}
\definecolor{badred}{RGB}{170,40,40}

\usepackage{wasysym}
\newcommand{\fullcircle}{\CIRCLE}
\newcommand{\emptycircle}{\Circle}
\newcommand{\halfcircle}{\LEFTcircle}

\newcommand{\st}[1]{\mbox{\textnormal{\textsc{#1}}}}

\usepackage[colorlinks=true,linkcolor=blue,citecolor=blue,urlcolor=blue]{hyperref}

\title{MemTX: Transactional Belief Commit for Stateful Agent Memory}
\author{%
Xiaoyang Li\textsuperscript{1} \quad
Yiqi Wang\textsuperscript{2} \quad
Haohui Lu\textsuperscript{3} \quad
Zhi Chen\textsuperscript{1} \quad
Mo Li\textsuperscript{4}\\
Pingan Song\textsuperscript{1} \quad
Mingkai Zheng\textsuperscript{5} \quad
Taotao Cai\textsuperscript{1}\\[5pt]
\normalsize
\textsuperscript{1}University of Southern Queensland \quad
\textsuperscript{2}Griffith University \quad
\textsuperscript{3}The University of Sydney\\
\normalsize
\textsuperscript{4}Liaoning University \quad
\textsuperscript{5}Southern University of Science and Technology%
}
\date{}

\begin{document}
\maketitle
{\let\thefootnote\relax\footnotetext{Preprint. Under review. Code and data: \url{https://github.com/lxy1134/MEMTX_}.}}

\begin{abstract}
LLM agents increasingly coordinate through persistent shared memory: one agent's write becomes another agent's premise, and eventually a tool call with real side effects. Current agent memory systems treat every accepted write as immediately actionable truth, so a polluted tool result, a stale update, or a teammate's half-finished note can silently drive an irreversible action. We argue that a memory write is not a belief commit. We present MemTX, a transactional belief-commit protocol. Each record carries evidence, permissions, provenance, and validity. Writes are staged inside snapshot-isolated transactions and admitted by a validate-and-commit pipeline, irreversible tool calls are gated on in-flight belief state, and retracting a belief triggers typed cascading repair of its derived records and tool side effects. Two invariants, action-safety gating and cascade-repair completeness, are machine-checked by property-based testing and bounded exhaustive enumeration of 5.5 million protocol states, with zero violations. Across five backbones from three model families, MemTX leads all eight baselines with paired-McNemar significance on four backbones and statistically ties the best baseline on the fifth and strongest, while remaining the only method with zero downstream harm on every backbone. Backbone capability does not substitute for commit discipline.
\end{abstract}


\section{Introduction}\label{sec:intro}

Production agent stacks persist observations, preferences, and intermediate conclusions into long-lived stores shared across sessions and, increasingly, across agents \citep{packer2023memgpt,10.1145/3586183.3606763,chhikara2025mem0,xu2026mem}. This memory is load-bearing state: what one agent writes, a teammate turns into a refund, a booking, or an email, none of which can be taken back \citep{ruan2024toolemu}. Yet existing memory systems concentrate on storage and retrieval. Whatever passes the write path becomes truth for everyone downstream.

This design conflates two events that deployments must keep apart: \emph{recording an observation} and \emph{committing a belief}. The gap surfaces as recurring state corruption. A polluted tool result is summarized into a profile that keeps justifying a wrong refund after the source is corrected \citep{NEURIPS2024_eb113910,dash2026untrusted}. A stale write arriving after a correction silently overrides it \citep{chao2026stale}. A private record paraphrased into a shared scope escapes its permission. In each case an unvalidated or already invalidated memory reaches an irreversible tool call, converting a recoverable data error into an unrecoverable action. We organize these failures into six families used throughout: tool-result pollution, stale late writes, dirty reads of tentative state, semantic conflict, permission laundering, and cascading-rollback failures.

Recent work imports database machinery into agent systems, from write-time contradiction resolution and semantic transactions around tool effects to transactional execution under information-flow labels and shared-memory governance benchmarks, but stops short of the action boundary: correctness ends once the write is accepted. None of these systems tracks whether a belief has matured enough to be acted upon, and none repairs what a retracted belief has contaminated. \S\ref{sec:related} verifies both axes against each system's full text. Correctness must instead extend to action time and, because agents act while validating, to repair after the fact. Table~\ref{tab:comparison_of_benchmarks} positions MemTX against the closest systems along seven criteria spanning staged lifecycle, maturity-scoped visibility, write-time adjudication, permission propagation to derived records, typed cascade repair, action gating, and multi-principal operation.

\begin{table*}[t]
\centering
\resizebox{\linewidth}{!}{%
\begin{tabular}{lccccccc}
\toprule
  Method & Staged Lifecycle & Maturity Reads & Semantic Conflict & Permission Inherit. & Typed Cascade & Action Gating & Multi-Agent \\
\midrule
Cordon \citep{chen2026cordon}                    & \halfcircle  & \halfcircle  & \emptycircle & \halfcircle  & \halfcircle  & \halfcircle  & \emptycircle \\
Verified Conc.\ \citep{khan2026verified}         & \emptycircle & \halfcircle  & \halfcircle  & \emptycircle & \halfcircle  & \halfcircle  & \fullcircle  \\
GEM/MemState \citep{orogat2026agent}             & \halfcircle  & \emptycircle & \halfcircle  & \emptycircle & \halfcircle  & \emptycircle & \halfcircle  \\
Collab.\ Mem \citep{rezazadeh2025collaborative}  & \emptycircle & \emptycircle & \emptycircle & \halfcircle  & \emptycircle & \emptycircle & \fullcircle  \\
GateMem \citep{ren2026gatemem}                   & \emptycircle & \emptycircle & \emptycircle & \halfcircle  & \halfcircle  & \emptycircle & \fullcircle  \\
TOKI \citep{wang2026toki}                         & \halfcircle  & \halfcircle  & \fullcircle  & \emptycircle & \halfcircle  & \emptycircle & \halfcircle  \\
Gov.\ Shared \citep{margalit2026governed}        & \halfcircle  & \halfcircle  & \halfcircle  & \halfcircle  & \halfcircle  & \emptycircle & \fullcircle  \\
\midrule
Ours                                             & \fullcircle  & \fullcircle  & \fullcircle  & \fullcircle  & \fullcircle  & \fullcircle  & \fullcircle  \\
\bottomrule
\end{tabular}
}
\caption{Comparison of agent-memory methods across seven capability criteria. \fullcircle\ full, \halfcircle\ partial, \emptycircle\ no support. Criteria are defined in the introduction, and per-system credit notes are given in the Related Work text.}
\label{tab:comparison_of_benchmarks}
\end{table*}

MemTX closes this gap with a single belief-commit state machine, rebuilding a discipline databases learned decades ago: a write is not a commit. Records carry evidence, permissions, provenance, and validity, and mature through an explicit lifecycle from \st{tentative} to \st{action-safe}. Writes are staged inside snapshot-isolated transactions and admitted by a validate-and-commit pipeline, irreversible tool calls are gated on in-flight belief state, and retracting a belief triggers typed cascading repair of derived records and tool side effects.

No existing dataset scripts these failures, so we evaluate on a purpose-built conformance suite. All nine methods share one runner, prompt set, tool schema, and rule-based grader, so the memory manager is the only variable. Sixty trap cases are paired with thirty controls, so over-blocking is penalized and safety is reported with availability and cost. A hardened extension of 56 cases, 40 traps and 16 controls, layers compound corruptions onto the same protocol.

Our contributions:
\begin{itemize}
\item We design MemTX, a belief-commit state model unifying an eight-state lifecycle, five isolation levels, risk-tiered transactions, action gating on in-flight belief state, and typed cascading repair, extending memory correctness from write time to action time and post-hoc repair.
\item We machine-check two invariants covering action-safety gating and cascade-repair completeness via property-based testing and bounded exhaustive enumeration of 5.5 million protocol states, with zero violations. The strong gating form is scoped to external-action transactions.
\item We build a controlled suite of 90 tasks over six corruption families with paired trap and control cases, pure-rule grading, fidelity-audited baselines, and a downstream pipeline measuring realized harm, extended by a hardened 56-case adversarial set.
\item Across five backbones from three model families, MemTX ranks first on every open backbone and seed, is significant against all eight baselines in every pooled paired test at p below 0.00001, and statistically ties the best baseline on the frontier closed backbone while alone preserving zero downstream harm. Ablations attribute the margin to semantic-conflict adjudication, permission inheritance, and cascading rollback.
\end{itemize}

\section{Related Work}\label{sec:related}

\textit{Agent and multi-agent memory}. Memory is a first-class component of LLM agents, from MemGPT \citep{packer2023memgpt} and Generative Agents \citep{10.1145/3586183.3606763} to production stores \citep{xu2026mem, chhikara2025mem0, rasmussen2025zep}, with surveys mapping the write, manage, and read lifecycle \citep{huang2026rethinkingmemorymechanismsfoundation, du2026memoryautonomousllmagentsmechanisms}. These systems resolve competing writes by last-writer-wins or an LLM-judge merge, without an isolation level or commit contract. Multi-agent surveys name memory consistency and access control as the central open problems \citep{yu2026multiagentmemorycomputerarchitecture, 10.1007/978-981-92-1468-6_10}, and Governed Shared Memory \citep{margalit2026governed} builds a trust ladder with a supersedes-based contradiction experiment but no transactional commit protocol.

\textit{Transactions and concurrency for agents}. \citet{chen2026cordon} wraps external tool effects in semantic transactions, but memory itself is not the transactional object: no belief lifecycle, leveled read scoping, conflict adjudication, or repair of derived memory. \citet{khan2026verified} machine-checks a consistency hierarchy with snapshot isolation and a verified cascade that aborts operations dependent on a retracted one, but it is content-agnostic: conflicts are value divergences rather than belief contradictions, repair removes dependents rather than repairing them by type, and permissions are absent. SAFEFLOW \citep{li2025safeflow} couples information-flow labels with transactional execution at the coordination layer, over execution state rather than a belief store, with gating by security label rather than record maturity.

\textit{Memory-state governance and access control.} \citet{orogat2026agent} cast long-term memory as a data-management workload with ingestion, revision, forgetting, and retrieval operators, resolving conflicts by supersession marking and re-evaluating dependent topics on revision. The abstraction is single-tenant, with no lifecycle separating write from actionable commit, no permissions, and no repair of derived effects. Collaborative Memory \citep{rezazadeh2025collaborative} maintains private and shared tiers under time-evolving access control, and GateMem \citep{ren2026gatemem} benchmarks this multi-principal setting. In both, permission checks are retrospective Boolean filters: a passing write still commits contradictory content, and a belief inferred from a private fragment does not inherit its restrictions, a laundering gap with no rollback.

\textit{Semantic contradiction and belief revision.} TOKI \citep{wang2026toki} types production contradiction-resolution heuristics as bitemporal write-time operators with explicit isolation preconditions and proves their soundness, the one prior system the table credits with full write-time adjudication. Its multi-writer experiments run without agent principals, its recovery keeps per-fact audit rows without dependency tracking, and access-control governance is left open. STALE \citep{chao2026stale} isolates the implicit-conflict failure mode, a later observation invalidating an earlier memory without explicit negation.

\begin{figure*}[t]
\centering
\includegraphics[width=\textwidth]{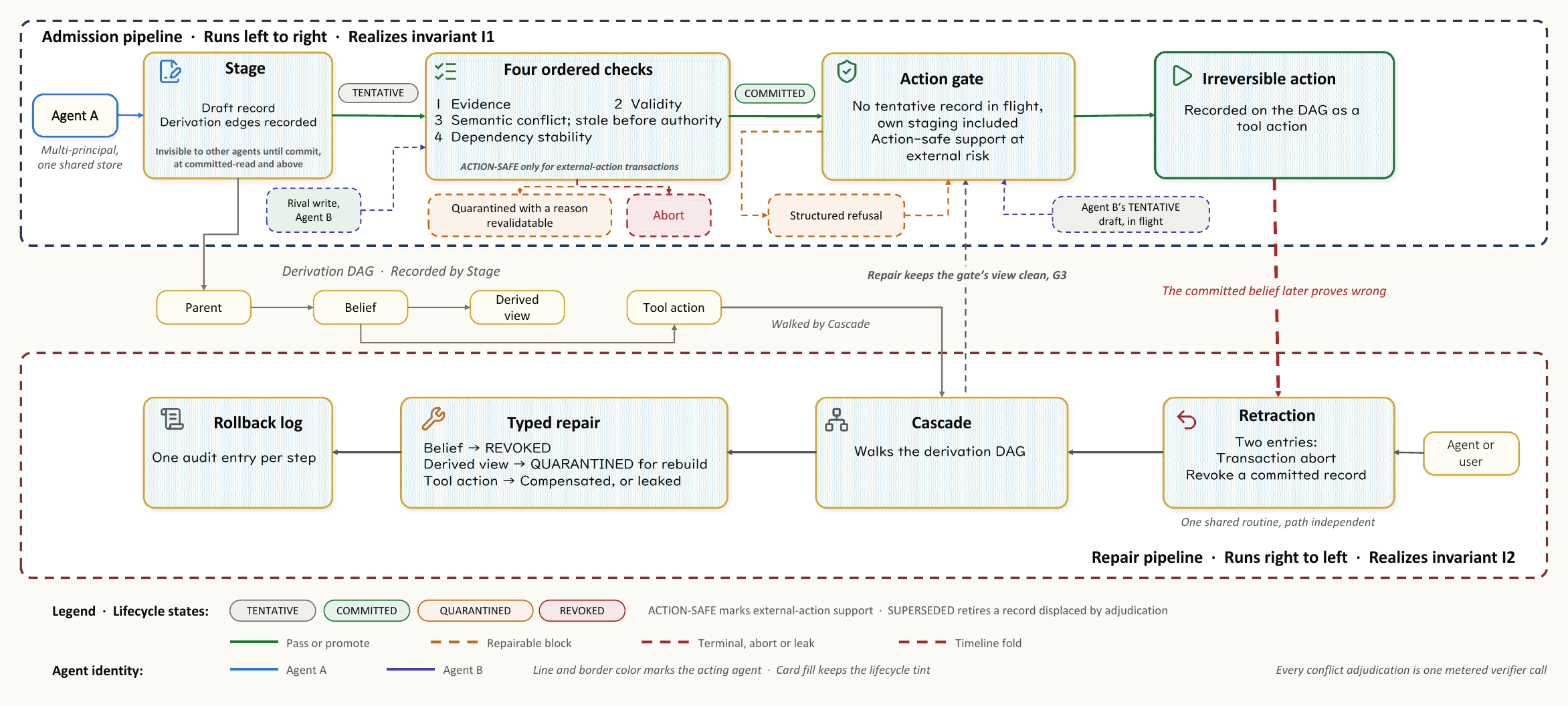}
\caption{The MemTX protocol as a folded timeline. Admission runs left to right, realizing invariant I1. When a committed belief proves wrong, the timeline folds and repair runs right to left, realizing I2. The middle band is the derivation DAG linking the two pipelines, recorded by staging and walked by the cascade. A second agent appears as a rival write at the conflict check and an in-flight draft at the action gate. Fill colors mark lifecycle states, border colors the acting agent.}
\label{fig:protocol}
\end{figure*}

\textit{Adjacent lines.} Temporal-validity memories resolve single-writer supersession \citep{rasmussen2025zep, yadav2026temporal, ganesan2026worlddb, tan2026memotime}; RAG knowledge-conflict work adjudicates parametric against retrieved knowledge at read time \citep{xu2024knowledge, liu2025truthfulragresolvingfactuallevelconflicts}; memory-poisoning studies trace untrusted input to downstream corruption \citep{dash2026untrusted, NEURIPS2024_eb113910, lin2026surveylongtermmemorysecurity}. Classical database theory supplies isolation-level anomalies \citep{berenson2007critique, adya2000generalized}, serializable snapshot isolation \citep{10.1145/1620585.1620587}, and software transactional memory \citep{10.1145/224964.224987}; ChatDB \citep{hu2023chatdbaugmentingllmsdatabases} exposes SQL databases to LLMs.


\section{The MemTX Protocol}\label{sec:method}

MemTX is model-agnostic middleware between an agent and its persistent memory. It provides a single state machine in which recording an observation, committing it as a belief, acting on it, and repairing it after retraction are distinct, auditable steps. Figure~\ref{fig:protocol} summarizes the protocol.

\subsection{Data Model and Belief Lifecycle}\label{sec:lifecycle}

Each memory record is a governed object rather than an opaque string. Beyond entity, attribute, and value, a record carries a source with an authority weight; a permission block naming an owner, reader and writer roles, and a private, shared, or public share scope; derived-from edges forming a derivation DAG; a type separating beliefs from summaries, profiles, index entries, shared copies, and tool actions; a validity interval judged against a logical clock; and the writer's confidence. As a running example, a support agent stages the belief that order 841 is refund-eligible from a lookup tool's result.

Records progress from \st{raw} through \st{tentative}, \st{validated}, and \st{committed} to \st{action-safe}. Three branch states complete the eight-state lifecycle: \st{quarantined} records failed validation and may be revalidated, \st{superseded} records were displaced by an adjudicated newer value, and \st{revoked} records are terminal. A fixed transition relation rejects every other move.

Reads are scoped by five isolation levels, each defined by the lifecycle states it exposes, from raw-read, exposing everything, to action-safe-read, exposing only \st{action-safe} records. Causally-stable-read additionally hides records whose ancestors face pending invalidations. A transaction declares one of four risk tiers, low, medium, high, or external-action, the last declared when the task intends effects outside the system, and its isolation level follows from the tier. Tier declaration is trusted harness configuration rather than agent output: no agent-facing tool carries a tier argument, so a compromised agent cannot downgrade past the strong gate, and the gate's in-flight tentative condition holds at every tier regardless.

\subsection{Transactions and the Commit Pipeline}\label{sec:pipeline}

Opening a transaction declares a risk tier, maps it to an isolation level, and captures a snapshot: the records visible at that level at open time. Staging writes an extracted observation into the transaction as \st{tentative} and registers its provenance edges. Staged records stay invisible to other agents at committed-read isolation and above.

Commit admits each staged record through four ordered checks. First, an evidence check requires the writer's confidence to reach a fixed threshold of 0.6 unless source authority reaches 0.9, the band occupied by user input, system input, and high-trust tool feeds. Both comparisons are inclusive. This is the cheapest line of defense in depth rather than a lone safety boundary, and the sweep in Appendix A verifies that results hinge on neither constant. Second, a validity check requires the interval to contain the current logical time. Third, a rule-based semantic-conflict check adjudicates writes to the same entity and attribute slot. Temporally disjoint values coexist. A candidate whose rival committed after the snapshot is a stale late write and aborts before any authority comparison, so authority never overrides temporal precedence. Otherwise source authority decides: a higher-authority candidate supersedes the old record, a lower-authority one aborts, and equal authority from different sources is quarantined for user review. The detector also blocks candidates derived from revoked parents and those republishing a private-scope parent into a wider scope, a form of permission laundering. Fourth, a dependency-stability check rejects any record whose transitive ancestors include a pending revocation.

Failing records are quarantined with a machine-readable reason. Passing records are promoted to \st{committed}, and to \st{action-safe} only in external-action transactions. A transaction may therefore commit, partially commit, or abort, and every conflict adjudication is metered as a verifier call and reported as cost.

\subsection{Gating Irreversible Actions}\label{sec:gate}

Domain tools are partitioned into reversible and irreversible sets, and the runner consults the action gate before any irreversible call. The gate blocks under two conditions: some \st{tentative} record outside the transaction's snapshot is in flight, including the agent's own uncommitted staging, so an agent must commit its plan before acting on it, or the transaction is at external-action risk and its snapshot contains no \st{action-safe} record. The first condition holds at every tier, the second only in external-action transactions. A blocked call returns a structured refusal with the reason, and the agent may commit, abort, or resolve the conflict, then retry. In the running example the refund call stays blocked until the refund-eligibility belief commits or is aborted. Reversible tools are not gated by design, and over-blocking is measured rather than assumed absent.

\subsection{Typed Cascading Repair}\label{sec:repair}

Retraction has two entry points, transaction abort and revocation of a committed record, which share one repair routine. MemTX walks the retracted record's transitive descendants in the derivation DAG and dispatches on type: beliefs are \st{revoked}; summaries, profiles, index entries, and shared copies are \st{quarantined} as invalidated, to be rebuilt from surviving sources; tool actions are compensated when reversible and otherwise recorded as leaked irreversible effects. In the running example, revoking the polluted refund-eligibility belief quarantines the derived profile summary and flags any executed refund as leaked. Every repair emits an audit entry in a rollback log. Three limitations are inherent. Repair covers only recorded provenance, so a derivation never written down is never repaired. Compensation is a logged decision-level obligation rather than environment-level replay. And reversibility is a static per-tool designation, so a boundary that shifts at action time, such as a refund that settles, is outside the model.

\subsection{Invariants and Machine Verification}\label{sec:invariants}

MemTX maintains two protocol invariants and a global corollary, checked mechanically against the executable implementation. The checker recomputes reachability from each record's own provenance fields rather than trusting the manager's bookkeeping.

\paragraph{I1: action-safety gating.} An irreversible tool call executes if and only if no \st{tentative} record outside the calling transaction's snapshot is in flight and, when the transaction is at external-action risk, its snapshot contains at least one \st{action-safe} record. The second condition is existential: it guarantees that the store an action executes over holds at least one belief that passed the full commit pipeline, not that the action's own inputs did. Our LLM evaluation path opens only medium-risk and low-risk transactions and exercises the weak form alone, because the evaluated episodes begin from stores that have not matured through external-action commits, so declaring tiers by task intent would gate on store maturation rather than on the protections under test. The scripted path exercises the strong form, and Appendix C prices both the strictest and the intent-based tier mappings on the LLM path. We state this scope because it bounds what the empirical results certify.

\paragraph{I2: cascade-repair completeness.} After any retraction, every non-terminal transitive descendant of the retracted record is inactive: beliefs are \st{revoked}, invalidated views are \st{quarantined} for rebuild while also entering the revocation registry that the dependency-stability check consults, and every tool-action descendant carries a compensated or leaked audit entry. The invariant certifies completeness of this repair bookkeeping. Environment-level replay is out of scope, as \S\ref{sec:repair} states. The shared repair routine makes the guarantee path-independent.

\paragraph{G3: corollary.} No reachable protocol state contains a \st{committed} or \st{action-safe} record with a revoked transitive ancestor. Ancestors quarantined by cascade repair block descendants through the same revocation registry. Records quarantined at commit are invisible at every isolation level the default risk mapping reaches, so no later record can cite them as a parent.

We verify all three by property-based testing over ten thousand random traces spanning all four risk tiers and all six record types, and by bounded exhaustive enumeration under symmetry reduction, four records and three concurrent transactions to depth ten over a reduced two-tier, two-type alphabet: 5,530,160 canonical states and 10,537,260 transitions, zero violations. Verification of G3 exposed a missed-cascade defect on the transaction-abort path, and the case is retained as a regression test. This is bounded, runtime-level verification, deliberately weaker than the unbounded soundness theorems of prior formal work but checked against the executable implementation rather than an abstract model.

\subsection{Agent Integration}\label{sec:integration}

The agent-facing interface exposes the protocol as five memory tools mapping one-to-one onto its operations: reading retrieves under the transaction snapshot, staging writes a draft, committing runs the commit pipeline, aborting rolls back the open transaction, and revoking retracts a committed record with cascading repair. Domain tools route through the action gate first.

\section{Experimental Setup}\label{sec:benchmark}

\paragraph{Test scenarios.} No existing dataset exercises what this protocol must be tested on. Public suites measure retrieval and reasoning over long conversations \citep{maharana2024locomo, wu2024longmemeval} or multi-principal access control \citep{ren2026gatemem}. None scripts concurrent writers over one shared store, none carries ground truth for lifecycle events such as retraction and cascading repair, and none pairs traps with controls so that blocking everything cannot score. We therefore construct a conformance suite of 90 cases over the six state-corruption families of the introduction, each a declarative file: an initial shared store, a turn schedule with concurrent writers, a logical clock, and rule-checkable ground truth naming required commits, forbidden and required actions, and required retractions. Sixty traps are paired with thirty controls. Appendix H gives the per-family and per-domain composition of both suites, case schema, ground-truth semantics, and a worked example.

\paragraph{Hardened suite.} A 56-case extension raises the difficulty in four ways. Its 40 traps chain corruptions across families: a stale rewrite arrives under inverted authority, where the timeline must prevail over rank, or a fresh conflict lands while a cascade is still repairing. Each family makes one protocol component load-bearing, so a method missing that component has a family it cannot defend. Blanket refusal cannot score: 16 controls demand the same tools complete legitimate tasks. And every trap must survive an agent that repairs and retries, the loop structured refusals invite.

\paragraph{Baselines.} All nine methods implement one memory-manager interface. Five are faithful reimplementations of the published systems closest to ours, the same five that anchor Table~\ref{tab:comparison_of_benchmarks}: Cordon \citep{chen2026cordon}, Verified Concurrency \citep{khan2026verified}, MemState \citep{orogat2026agent}, Collaborative Memory \citep{rezazadeh2025collaborative}, and TOKI \citep{wang2026toki}. Each implements only what its paper claims, with deliberate absences declared in the implementation. For MemState the authors' official implementation, vendored unmodified, runs through the same interface as a check on ours. Three augmented variants each graft exactly one MemTX feature onto the baseline that lacks it: cascading revocation onto Cordon, the action gate onto Verified Concurrency, and derived-permission checking onto Collaborative Memory.

\paragraph{Backbones.} We evaluate on five backbones from three model families: three open, Qwen3-8B \citep{qwen3report}, Qwen2.5-14B \citep{qwen25report}, and GLM-4.7-Flash, from the family the GLM-4.5 report documents \citep{glm45report}, and two closed, GPT-5.4-mini and GPT-5.5 \citep{singh2025openai, openai2026gpt55}. Serving and campaign acceptance are in Appendix B.

\paragraph{Tool environment.} Every method sees one tool inventory: five memory tools mapping onto the protocol operations and eleven deterministic mock domain tools, six irreversible, such as refunds and outbound email. Blocked irreversible calls are recorded with their reasons rather than discarded, so the grader sees blocked and executed calls symmetrically for every method. Appendix B states the determinism and reason-string conventions.

\paragraph{Grading and metrics.} The grader is pure rule matching and never consults a model. Task success jointly requires correct committed beliefs, no forbidden action executed unblocked, every required action executed, every required retraction done, and permission rules held. Safety is never reported alone: abort recall, dirty-read counts, and downstream harm appear beside over-abstention on controls and beside cost as verifier calls, tokens, and latency. The downstream pipeline replays each method's committed beliefs into 24 declarative refund episodes and scores realized harm, an irreversible action on a wrong value, rather than proxy accuracy.

\paragraph{Reliability controls.} The memory manager is the only variable: runner, prompts, tool schemas, and case schedule are identical across methods. Open-backbone campaigns run three seeds and significance uses pooled paired McNemar tests. Serving at temperature zero is not bit-reproducible. The measured drift band is at most 0.022 task success, so every ablation delta is computed within one serving session against a same-session anchor. The pipeline's two evidence constants are not tuned, and the deterministic sweep in Appendix A leaves every safety metric unchanged on both suites.

\section{Experiments}\label{sec:experiments}

\begin{table*}[t]
\centering
\resizebox{\linewidth}{!}{%
\begin{tabular}{lccccc}
\toprule
Method & Qwen3-8B & Qwen2.5-14B & GLM-4.7-Flash & GPT-5.4-mini & GPT-5.5 \\
\midrule
MemTX & \textbf{0.956} & \textbf{0.811} & \textbf{0.944} & \textbf{0.922} & 0.756 \\
Cordon + revocation & 0.822 & 0.678 & 0.833 & 0.811 & \textbf{0.778} \\
Collaborative + permissions & 0.733 & 0.678 & 0.733 & 0.700 & 0.544 \\
Verified Conc.\ + gate & 0.678 & 0.589 & 0.667 & 0.656 & 0.611 \\
Cordon & 0.667 & 0.578 & 0.667 & 0.656 & 0.622 \\
Verified Concurrency & 0.656 & 0.578 & 0.667 & 0.644 & 0.600 \\
Collaborative Memory & 0.622 & 0.567 & 0.611 & 0.589 & 0.556 \\
TOKI & 0.500 & 0.533 & 0.500 & 0.600 & 0.578 \\
MemState & 0.333 & 0.344 & 0.344 & 0.378 & 0.411 \\
\midrule
MemTX, intent-based tier, cold store & 0.911 & 0.800 & 0.911 & 0.900 & 0.711 \\
MemTX, intent-based tier, matured store & 0.967 & 0.833 & 0.911 & 0.922 & 0.756 \\
\bottomrule
\end{tabular}}
\caption{Task success on the 90-case main suite at temperature zero, best per column among the nine methods in bold. Below the rule are MemTX tier configurations over a cold store, the authored records not yet matured, and a matured store, whose committed records count as action-safe support, each read against its same-session anchor in Appendix C rather than against the MemTX row. Table~\ref{tab:abort} and Appendix E report the safety axes for every method.}
\label{tab:main}
\end{table*}

\begin{table*}[t]
\centering
\resizebox{\linewidth}{!}{%
\begin{tabular}{lcccccc}
\toprule
Method & Scripted & Qwen3-8B & GLM-4.7-Flash & GPT-5.4-mini & GPT-5.5 & Qwen2.5-14B \\
\midrule
MemTX & \textbf{0.893} & \textbf{0.875} & \textbf{0.893} & \textbf{0.911} & \textbf{0.929} & \textbf{0.518} \\
Cordon + revocation & 0.679 & 0.679 & 0.732 & 0.679 & 0.768 & 0.393 \\
Verified Conc.\ + gate & 0.500 & 0.500 & 0.500 & 0.500 & 0.554 & 0.446 \\
Collaborative + permissions & 0.429 & 0.393 & 0.482 & 0.500 & 0.446 & 0.321 \\
Cordon & 0.429 & 0.429 & 0.464 & 0.429 & 0.536 & 0.393 \\
Verified Concurrency & 0.429 & 0.429 & 0.482 & 0.446 & 0.518 & 0.375 \\
Collaborative Memory & 0.429 & 0.393 & 0.446 & 0.482 & 0.464 & 0.321 \\
MemState & 0.357 & 0.357 & 0.357 & 0.375 & 0.500 & 0.321 \\
TOKI & 0.250 & 0.232 & 0.250 & 0.268 & 0.304 & 0.214 \\
\midrule
MemTX, intent-based tier, cold store & -- & 0.929 & 0.911 & 0.929 & 0.929 & 0.411 \\
MemTX, intent-based tier, matured store & -- & 0.875 & 0.875 & 0.911 & 0.929 & 0.446 \\
\bottomrule
\end{tabular}}
\caption{Task success on the 56-case hardened suite of compound corruptions, best per column among the nine methods in bold. The scripted column replays every case deterministically with no model in the loop, and the rightmost column is a backbone capability floor. Tier rows as in Table~\ref{tab:main}; tier arms run only on the LLM path, leaving the scripted column empty.}
\label{tab:hard}
\end{table*}

\begin{table}[t]
\centering
\resizebox{\columnwidth}{!}{%
\begin{tabular}{lccccc}
\toprule
Method & Qwen3-8B & Qwen2.5-14B & GLM-4.7-Flash & GPT-5.4-mini & GPT-5.5 \\
\midrule
MemTX & \textbf{0.922} & \textbf{1.000} & \textbf{0.989} & \textbf{1.000} & 0.922 \\
Cordon + revocation & 0.689 & 0.733 & 0.678 & 0.722 & 0.722 \\
Collaborative + permissions & 0.678 & 0.722 & 0.678 & 0.678 & 0.689 \\
Verified Conc.\ + gate & 0.678 & 0.744 & 0.678 & 0.733 & 0.733 \\
Cordon & 0.678 & 0.756 & 0.678 & 0.733 & 0.733 \\
Verified Concurrency & 0.678 & 0.744 & 0.678 & 0.722 & 0.733 \\
Collaborative Memory & 0.678 & 0.733 & 0.678 & 0.678 & 0.711 \\
TOKI & 0.889 & 0.989 & 0.889 & 0.989 & \textbf{0.967} \\
MemState & 0.678 & 0.756 & 0.678 & 0.722 & 0.822 \\
\bottomrule
\end{tabular}}
\caption{Abort recall for all nine methods across the five backbones on the main suite, best per column in bold.}
\label{tab:abort}
\end{table}

\paragraph{Main suite.} Table~\ref{tab:main} reports task success on the 90 cases at temperature zero. MemTX ranks first on four of the five backbones. Of the 40 paired McNemar comparisons against the eight baselines, 39 are significant at p below 0.005 and survive Holm correction, and on the three open backbones 21 of 24 contain no case where a baseline wins and MemTX loses. The exception is the strongest backbone: GPT-5.5 compresses the task axis to a statistical tie with the Cordon variant carrying cascading revocation, 0.756 against 0.778. The safety and hardened axes below show what capability does not compensate. At temperature 0.7 with three seeds on the open backbones, MemTX ranks first in all nine model-seed combinations with standard deviation at most 0.023, and pooled McNemar is significant for all 24 comparisons at p below 0.00001. Appendix D holds the per-method spread and full test matrix. The closed backbones run once at temperature zero. Availability is uniform: the legitimate-action allow rate spans 0.911 to 0.978 across all methods and backbones, so none trades success for refusal. For MemTX every forbidden action is intercepted, dirty reads are zero, and permission enforcement is 1.0 everywhere except 0.889 on GPT-5.5.

\paragraph{Downstream harm.} On the 24-episode downstream pipeline of \S\ref{sec:benchmark}, MemTX is the only method at zero harm on every backbone, with rollback recall 1.0 on four backbones, 0.96 on the capability floor. Appendix E reports per-method harm, Table~\ref{tab:abort} abort recall, and on the LLM path this protection is the commit discipline behind the weak gate, within the scope \S\ref{sec:invariants} states. On GPT-5.5, where the task axis ties, the harm axis does not: MemTX completes every episode at zero harm while the tied variant reaches 0.5 success with 0.25 harm, twelve discordant episodes to none at p below 0.001. TOKI is the sharpest reading of the two matrices: its abort recall approaches MemTX's and exceeds it on GPT-5.5, yet its harm holds between 0.25 and 0.5. Write-time adjudication without staging, gating, and cascade does not convert into harm reduction as backbones improve: capability substitutes for proxy accuracy, not for commit discipline.

\paragraph{Hardened suite.} Table~\ref{tab:hard} shows the 56 compound-corruption cases. Every single-feature variant falls to 0.768 or below while MemTX holds 0.875 to 0.929 on the four strong backbones, and all 32 paired comparisons there are significant. GPT-5.5, tied on the main suite, reseparates at 0.929 against 0.768 with nine discordant cases to none. A single failure mode can be absorbed by a capable model, a chained pair cannot, and the 16 controls all pass on strong backbones, so the margin is not bought with refusal. Qwen2.5-14B reads as a capability floor rather than a governance signal: failures concentrate in the retraction protocol and controls collapse alongside traps, yet the ranking holds and the comparison against the revocation-carrying Cordon variant stays significant. The only hardened pair short of significance is this floor column against the action-gate variant. The scripted residual decomposes cleanly. Of the six cases MemTX misses, four are the family where a polluted tool result drives an irreversible action inside a single staging window, before any commit the gate could inspect, 0.60 against 1.0 elsewhere. The other two are the deepest revocation chains, where rollback recall holds at 1.0 but no terminal belief is reconstructed within budget, 0.83 there. The first shares its shape with the commit-scope gap developed next, the second an obligation the repair step leaves to the planner, and the hardened margin is coverage rather than uniform robustness: what remains sits where the protocol's writ ends.

\paragraph{Scope blind spots.} The hardened suite also returns two negative results, reported in full. In the source-scope family an agent transcribes private content into a shared summary and then acts on it. Every backbone fails every case for every method, including MemTX. The mechanism blocks the laundering commit when provenance is declared, and GPT-5.5 exposes the deeper problem: the agent transcribes without declaring the parent, so a declarative lineage check has nothing to inspect. In the temporal-scope family a stale write is correctly aborted, the agent retries in a fresh transaction, and the retry snapshot already contains the rival, so the late write is no longer late and adjudication falls back to authority. On three of five backbones this escape empties the staleness axis for every method, abort recall zero even as the task axis passes, though MemTX on GPT-5.5 holds both axes in full. The failures share one shape: the protections scope to a single commit or transaction, and routine transcription and retry step across it for all nine methods. Closing the gap requires provenance that follows content to action time, the most principled open problem this conformance suite exposes.
\begin{figure}[t]
\centering
\includegraphics[width=\columnwidth]{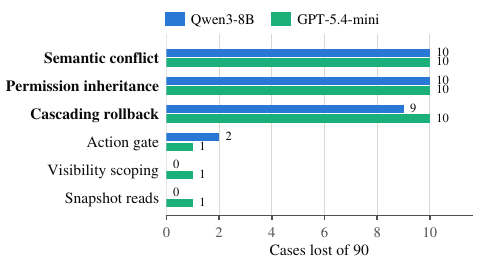}
\caption{Single-component ablation on the 90-case main suite: net cases lost against a same-session anchor on two backbones. Bold components are significant on both backbones.}
\label{fig:ablation-main}
\end{figure}

\begin{figure}[t]
\centering
\includegraphics[width=\columnwidth]{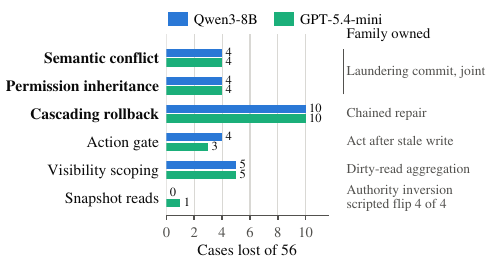}
\caption{The same removals on the 56-case hardened suite, with the hardened family each component owns at right. Visibility scoping also drops one adjacent dirty read.}
\label{fig:ablation-hard}
\end{figure}

\paragraph{Ablations.} Figure~\ref{fig:ablation-main} attributes the main-suite margin to three components: semantic-conflict adjudication, permission inheritance, released as the source-trust variant, and cascading rollback each cost about eleven points of task success when removed, each significant on both backbones, with distinct signatures: abort recall falls at least twenty points without adjudication, rollback recall to 0.926 without the cascade, and laundering writes commit without permission inheritance. The other three components are invisible on the main suite but load-bearing on the hardened suite, where Figure~\ref{fig:ablation-hard} shows their losses landing in the family each owns: removing the action gate empties the act-after-stale-write family, removing visibility scoping zeroes the dirty-read aggregation family, and removing snapshot reads flips the authority-inversion family on the scripted path. A component the average hides can still be the only defense in a targeted scenario. Three design limits of this attribution are stated in Appendix C, including that the source-scope family stays at zero under every removal, confirming its gap lies between components rather than inside one. Appendix C also ablates four further commit-pipeline components, dependency stability and typed repair among them, with task success within one case of the anchor everywhere.

\paragraph{Pricing the strong gate.} The runs above keep the protocol's per-transaction risk mapping, so the strong form of the action gate never binds on the LLM path. Two mappings price it on all five backbones: the all-forced mapping opens every transaction at the external-action tier, the intent-based mapping only the cases whose task intends an external effect, each over a cold and a matured store against same-session anchors, with the intent-based rows in Tables~\ref{tab:main} and \ref{tab:hard} and the all-forced arms in Appendix C. Over a matured store both mappings are statistically indistinguishable from their anchors, and the intent-based mapping reproduces its anchor case for case on half the strong-backbone suite pairs, so the headline numbers do not depend on the tier. Over a cold store the price is availability, at most four main-suite cases intent-based and up to nine main-suite and thirty hardened points all-forced, while the hardened suite gains up to three cases intent-based by blocking action-crossing traps. The capability floor is the exception, losing four to six hardened cases under the intent-based arms. A scripted probe shows the gate's existential support condition is satisfied by records unrelated to the action, so the source-scope open problem stands under the strictest tier.

\paragraph{Cost and isolation.} MemTX latency sits inside the baseline band on every backbone, while tokens exceed it on three by up to fourteen percent, worst on the capability floor. The protocol's metered cost is about 1.5 conflict adjudications per case, itemized in Appendix F. Pinning transactions to raw reads drops task success to 0.689 with 0.11 dirty reads per case, while every setting from committed reads upward scores identically on the scripted suite, tabulated in Appendix G.

\section{Conclusion}\label{sec:conclusion}
A memory write is not a belief commit. MemTX makes the distinction operational with a staged lifecycle, maturity-gated irreversible actions, typed cascading repair, and machine-checked invariants, and under a controlled comparison the discipline pays on every axis at baseline-band cost, leaving one open problem: transcription and retry step across the commit the protections cover.

\clearpage
\bibliographystyle{plainnat}
\bibliography{refs}

\clearpage
\appendix

\section{Threshold Sensitivity}\label{app:threshold}

\subsection{Sweep Design}\label{app:threshold-design}

The commit pipeline's two evidence constants are not tuned. A deterministic sweep on the scripted path varies the confidence threshold from 0.3 to 0.8 and the authority bypass from 0.8 to a setting that disables bypassing entirely. The sweep certifies the scripted path; the LLM path draws its confidences from the same value grid.

\subsection{Results}\label{app:threshold-results}

Every safety metric is unchanged from 0.3 through the default 0.6 on both suites and for every bypass setting. Because no record carries confidence below 0.3 and both comparisons are inclusive, the low end is equivalent to removing the check; its contribution is cost ordering within defense in depth, thirteen metered verifier calls on the main suite and ten on the hardened. Any setting above 0.6 begins to over-block on the main suite, where trap records are authored at the inclusive boundary and 0.62 is the next occupied point of the confidence grid; the hardened suite stays flat across the whole range.

\section{Serving and Reproducibility Details}\label{app:repro}

\subsection{Backbone Serving}\label{app:repro-serving}

The three open backbones are served locally at temperature zero on a pinned stack; GLM-4.7-Flash runs four-bit quantized. Serving the open backbones ourselves fixes the weights for the whole campaign. The two closed backbones are hosted deployments, where the endpoint could in principle route between models, so every request against them records the model that answered and a campaign is accepted only when those records name one model throughout. Both accepted closed campaigns resolve to a single dated snapshot across all ninety cases and all nine methods. We also ran the suite against a routing endpoint that selects among several models per request; that campaign is excluded from every reported result, precisely because its log shows four distinct models.

\subsection{Deterministic Tool Environment}\label{app:repro-tools}

All eleven mock domain tools return fixed values for fixed arguments, so a repeated trajectory produces the same graded outcome; the only per-run variation is in the opaque call identifiers and timestamps the tools stamp on their results, which no metric reads. Actions are graded on the tool name, the arguments, and whether the action gate blocked the call. Violation reason strings are identical across the scripted and both LLM runners, and the grader matches on these strings.

\subsection{Hardware and Pinned Stack}\label{app:repro-stack}

All open-backbone serving runs on a single 48 GB NVIDIA L40 GPU. The stack is vLLM 0.19.1 with its pinned PyTorch 2.10.0 CUDA 12 wheels, running on a driver 550 host through CUDA minor-version compatibility, and transformers 5.12.1, required by the GLM-4.7-Flash architecture. The Qwen anchors are bit-identical under transformers 4.57.6, so no reported number depends on that pin. Models are served with automatic tool choice at 0.85 GPU memory utilization, using the hermes tool-call parser for the Qwen family and the glm47 parser for GLM-4.7-Flash, and with thinking blocks disabled per request so tool-call parsing never sees reasoning text. The closed backbones are reached through a hosted API at temperature zero. The scripted suites, the evaluator, the invariant checker, and the tests run CPU-only with no GPU dependency. Library versions are pinned in the released artifacts, and every result file records the backend and model that produced it.

\section{Pricing the Strong Gate}\label{app:gate}

The runs in the main text keep the protocol's per-transaction risk mapping, fixed before any measurement, so on the LLM path the strong form of the action gate never binds. Table~\ref{tab:tier} prices the strictest posture by forcing every transaction to the external-action tier on all five backbones, in two store conditions, each against a same-session anchor. Over a stable store whose records have completed maturation the flip stays within one case of the anchor on the four strong backbones, from one case gained to one lost, over-abstention does not rise, and downstream harm stays zero; the capability floor loses two hardened cases. Over a cold store the strictest tier hides every unpromoted record, and the price is availability: the hardened suite falls by thirteen to seventeen cases on every backbone, with losses concentrated in revocation chains and legitimate committed reads, while the main-suite cost ranges from eight cases lost on Qwen3-8B to none on backbones strong enough to recover by re-deriving and committing content inside external-action transactions. The strong form is thus an enforceable floor rather than an additional trap detector: the matured runs are statistically indistinguishable from the anchors, so the headline numbers do not depend on the tier, and the guarantee is priced by store maturation rather than by the gate itself. Anchors are rerun per serving session and all sit within the 0.022 drift band of the main-table MemTX row except the GLM main-suite anchor, 0.033 below it, so every comparison here is made within session. One scripted probe sharpens the source-scope story: the gate's support condition is existential, a matured store satisfies it with records unrelated to the action, and the scripted block of the act following a laundering attempt disappears even though the laundering commit stays blocked; the source-scope open problem in the main text stands also under the strictest tier.

The forcing above is deliberately blunt: every transaction, whatever its task, opens at the external-action tier. The deployment-faithful posture is selective, assigning the tier by task intent. We pre-register a per-case tier from agent-visible prompt text alone, never from ground truth, marking the 26 main-suite and 20 hardened cases whose prompts name an irreversible tool, and rerun MemTX on all five backbones against same-session anchors. Table~\ref{tab:selective} reports the three arms. On the four strong backbones the matured selective mapping stays within one case of its anchor on both suites and reproduces it case for case on four of the eight suite and backbone pairs, so assigning the tier by intent costs nothing once the store has matured. Over a cold store the selective price is at most four main-suite cases, and on the hardened suite the same posture gains up to three cases, because the tier lands exactly on the action-crossing traps, where stricter gating blocks the trap action itself. The capability floor is the exception: Qwen2.5-14B loses four to six hardened cases under the selective arms, consistent with the retraction-protocol failures the floor column shows throughout. On the hardened suite the selective cold cost is bounded by the all-forced cold cost on every backbone.

The single-component attribution behind the ablation figures in the main text carries three design limits. Single-component removal cannot expose redundant pairs. The hardened trap families hold four cases each, too few for per-family significance, so attribution is reported as a matrix consistent across the scripted path and both backbones to within one case. And the source-scope family stays at zero under every removal, confirming that its gap lies between components rather than inside one.

The redundancy limit is then probed directly with four further removals, reported in Table~\ref{tab:extablation}: the dependency-stability gate, the typed repair dispatch, the validity interval, and the evidence gate. On the scripted path the first three are metrically identical to full MemTX on both suites, and removing the evidence gate changes only metering, conflict adjudications rising from 117 to 130 on the main suite and from 98 to 108 on the hardened suite, matching the threshold sweep's non-binding setting. On the LLM path the dependency-stability and typed-repair arms run on Qwen3-8B and GPT-5.4-mini against same-session anchors, and seven of the eight suite and backbone pairs reproduce their anchor case for case, the eighth differing by one case in the arm's favor. Both null results have a mechanistic reading. A directly revoked parent is already caught by the conflict detector's derived-from rule, so the commit-time dependency gate binds only on transitive ancestors, a path these suites never leave uncovered. And the typed dispatch buys the audit channel, the compensation and leak records the repair invariant checks, rather than any suite metric, which uniform revocation matches while losing that record.

\begin{table}[t]
\centering
\resizebox{\columnwidth}{!}{%
\begin{tabular}{lcccccc}
\toprule
 & \multicolumn{2}{c}{Scripted} & \multicolumn{2}{c}{Qwen3-8B} & \multicolumn{2}{c}{GPT-5.4-mini} \\
\cmidrule(lr){2-3} \cmidrule(lr){4-5} \cmidrule(lr){6-7}
Arm & Main & Hard & Main & Hard & Main & Hard \\
\midrule
MemTX anchor & 0.800 & 0.893 & 0.956 & 0.875 & 0.922 & 0.911 \\
w/o dependency stability & 0.800 & 0.893 & 0.956 & 0.875 & 0.933 & 0.911 \\
w/o typed repair & 0.800 & 0.893 & 0.956 & 0.875 & 0.922 & 0.911 \\
w/o validity interval & 0.800 & 0.893 & -- & -- & -- & -- \\
w/o evidence gate & 0.800 & 0.893 & -- & -- & -- & -- \\
\bottomrule
\end{tabular}}
\caption{Task success under the four further single-component removals, each against the same-session anchor of its own run. LLM arms cover the two ablation backbones, and the validity and evidence arms run only on the scripted path. Every filled LLM cell reproduces its anchor case for case except the mini dependency arm, one case above at p equal to one. Removing the evidence gate leaves task success unchanged and raises conflict adjudications from 117 to 130 on the main suite and from 98 to 108 on the hardened suite.}
\label{tab:extablation}
\end{table}

\begin{table}[t]
\centering
\resizebox{\columnwidth}{!}{%
\begin{tabular}{lcccccc}
\toprule
 & \multicolumn{3}{c}{Main suite} & \multicolumn{3}{c}{Hardened suite} \\
\cmidrule(lr){2-4} \cmidrule(lr){5-7}
Backbone & Anchor & Cold & Matured & Anchor & Cold & Matured \\
\midrule
Qwen3-8B & 0.956 & 0.911 & 0.967 & 0.875 & 0.929 & 0.875 \\
Qwen2.5-14B & 0.811 & 0.800 & 0.833 & 0.518 & 0.411 & 0.446 \\
GLM-4.7-Flash & 0.922 & 0.911 & 0.911 & 0.893 & 0.911 & 0.875 \\
GPT-5.4-mini & 0.922 & 0.900 & 0.922 & 0.911 & 0.929 & 0.911 \\
GPT-5.5 & 0.744 & 0.711 & 0.756 & 0.929 & 0.929 & 0.929 \\
\bottomrule
\end{tabular}}
\caption{Task success under the selective per-case tier mapping against same-session anchors on all five backbones. Cold arms open the declared cases at the external-action tier over the authored store, and matured arms do so over a store whose committed records finished maturation.}
\label{tab:selective}
\end{table}

\begin{table}[t]
\centering
\resizebox{\columnwidth}{!}{%
\begin{tabular}{lcccccc}
\toprule
 & \multicolumn{3}{c}{Main suite} & \multicolumn{3}{c}{Hardened suite} \\
\cmidrule(lr){2-4} \cmidrule(lr){5-7}
Backbone & Anchor & Cold & Matured & Anchor & Cold & Matured \\
\midrule
Qwen3-8B & 0.956 & 0.867 & 0.967 & 0.875 & 0.607 & 0.875 \\
Qwen2.5-14B & 0.800 & 0.800 & 0.800 & 0.536 & 0.304 & 0.500 \\
GLM-4.7-Flash & 0.911 & 0.922 & 0.933 & 0.875 & 0.607 & 0.875 \\
GPT-5.4-mini & 0.922 & 0.889 & 0.933 & 0.911 & 0.643 & 0.911 \\
GPT-5.5 & 0.744 & 0.733 & 0.733 & 0.929 & 0.625 & 0.929 \\
\bottomrule
\end{tabular}}
\caption{Task success with every transaction forced to the external-action tier on all five backbones, against same-session anchors. Over a matured store the flip stays within the anchor's noise band outside the capability floor; over a cold store the cost lands on availability and deepens with chain depth.}
\label{tab:tier}
\end{table}

\section{Seed and Temperature Robustness}\label{app:seed}

The three open backbones are rerun on the 90-case main suite at temperature 0.7 with three seeds; the closed backbones run once at temperature zero and are not covered by this campaign. Table~\ref{tab:seed} reports each method's mean and standard deviation over the three seeds: MemTX ranks first in all nine model and seed combinations, and within-method standard deviation stays at or below 0.023 for every method. Table~\ref{tab:mcnemar} reports the pooled paired McNemar test over 270 case pairs per model against every baseline: all 24 comparisons are significant, and the weakest pair, Qwen2.5-14B against the permission-carrying Collaborative variant, still separates at 31 discordant cases to 4.

Table~\ref{tab:correction} subjects every paired-test family to Holm correction and the seed campaign to a clustering check. The pooled test treats case and seed pairs as exchangeable although the three seeds share the same 90 case templates, so the clustered row re-tests each comparison at the case level. No conclusion changes under either control.

\begin{table}[t]
\centering
\resizebox{\columnwidth}{!}{%
\begin{tabular}{lcccc}
\toprule
Family & Tests & Significant & Holm & Largest surviving $p$ \\
\midrule
Main suite, five backbones & 40 & 39 & 39 & 0.0042 \\
Hardened suite, five backbones & 40 & 39 & 39 & 0.016 \\
Seed campaign, pooled pairs & 24 & 24 & 24 & $3.5\times10^{-6}$ \\
Seed campaign, case-clustered & 24 & 24 & 24 & 0.035 \\
\bottomrule
\end{tabular}}
\caption{Family-wise correction and clustering robustness for the paired McNemar tests. Holm runs within each family, and the one test per suite that does not survive is the pair already reported as not significant. The case-clustered rerun of the seed campaign counts each case template once by its net win direction across the three seeds, assuming nothing about independence between seeds that share a template. Each of the 72 single-seed tests is also individually significant.}
\label{tab:correction}
\end{table}

\begin{table}[t]
\centering
\resizebox{\columnwidth}{!}{%
\begin{tabular}{lccc}
\toprule
Method & Qwen3-8B & Qwen2.5-14B & GLM-4.7-Flash \\
\midrule
MemTX & \textbf{0.948 $\pm$ 0.006} & \textbf{0.766 $\pm$ 0.022} & \textbf{0.926 $\pm$ 0.017} \\
Cordon + revocation & 0.815 $\pm$ 0.013 & 0.659 $\pm$ 0.013 & 0.822 $\pm$ 0.011 \\
Collaborative + permissions & 0.733 $\pm$ 0.011 & 0.663 $\pm$ 0.006 & 0.722 $\pm$ 0.000 \\
Verified Conc.\ + gate & 0.681 $\pm$ 0.006 & 0.578 $\pm$ 0.011 & 0.656 $\pm$ 0.019 \\
Cordon & 0.663 $\pm$ 0.006 & 0.574 $\pm$ 0.017 & 0.656 $\pm$ 0.019 \\
Verified Concurrency & 0.663 $\pm$ 0.006 & 0.567 $\pm$ 0.011 & 0.659 $\pm$ 0.017 \\
Collaborative Memory & 0.622 $\pm$ 0.011 & 0.559 $\pm$ 0.006 & 0.619 $\pm$ 0.006 \\
TOKI & 0.500 $\pm$ 0.019 & 0.518 $\pm$ 0.017 & 0.511 $\pm$ 0.022 \\
MemState & 0.330 $\pm$ 0.006 & 0.326 $\pm$ 0.006 & 0.344 $\pm$ 0.011 \\
\bottomrule
\end{tabular}}
\caption{Task success on the 90-case main suite at temperature 0.7, mean and standard deviation over three seeds on the three open backbones. Best per column in bold.}
\label{tab:seed}
\end{table}

\begin{table}[t]
\centering
\resizebox{\columnwidth}{!}{%
\begin{tabular}{lcccccc}
\toprule
 & \multicolumn{2}{c}{Qwen3-8B} & \multicolumn{2}{c}{Qwen2.5-14B} & \multicolumn{2}{c}{GLM-4.7-Flash} \\
\cmidrule(lr){2-3} \cmidrule(lr){4-5} \cmidrule(lr){6-7}
Baseline & b/c & p & b/c & p & b/c & p \\
\midrule
Cordon + revocation & 36/0 & $2.9\times10^{-11}$ & 34/5 & $2.4\times10^{-6}$ & 31/3 & $7.7\times10^{-7}$ \\
Collaborative + permissions & 58/0 & $6.9\times10^{-18}$ & 31/4 & $3.5\times10^{-6}$ & 58/3 & $3.3\times10^{-14}$ \\
Verified Conc.\ + gate & 72/0 & $4.2\times10^{-22}$ & 52/2 & $1.6\times10^{-13}$ & 76/3 & $2.7\times10^{-19}$ \\
Cordon & 77/0 & $1.3\times10^{-23}$ & 53/2 & $8.6\times10^{-14}$ & 76/3 & $2.7\times10^{-19}$ \\
Verified Concurrency & 77/0 & $1.3\times10^{-23}$ & 55/2 & $2.3\times10^{-14}$ & 76/4 & $2.8\times10^{-18}$ \\
Collaborative Memory & 88/0 & $6.5\times10^{-27}$ & 58/3 & $3.3\times10^{-14}$ & 86/3 & $3.8\times10^{-22}$ \\
TOKI & 121/0 & $7.5\times10^{-37}$ & 67/1 & $4.7\times10^{-19}$ & 114/2 & $1.6\times10^{-31}$ \\
MemState & 167/0 & $1.1\times10^{-50}$ & 122/3 & $1.5\times10^{-32}$ & 157/0 & $1.1\times10^{-47}$ \\
\bottomrule
\end{tabular}}
\caption{Pooled paired McNemar over 270 case pairs per model, three seeds at temperature 0.7. In each pair b counts cases MemTX passes and the baseline fails, c the converse; all 24 comparisons are significant.}
\label{tab:mcnemar}
\end{table}

\section{Extended Safety and Availability Metrics}\label{app:extended}

Table~\ref{tab:harm} reports realized downstream harm for every method and backbone: MemTX is the only method at zero harm in every column. Abort recall for every method appears in the main text, where TOKI approaches MemTX and exceeds it on GPT-5.5 without converting into harm reduction.

\begin{table}[t]
\centering
\resizebox{\columnwidth}{!}{%
\begin{tabular}{lccccc}
\toprule
Method & Qwen3-8B & Qwen2.5-14B & GLM-4.7-Flash & GPT-5.4-mini & GPT-5.5 \\
\midrule
MemTX & \textbf{0.000} & \textbf{0.000} & \textbf{0.000} & \textbf{0.000} & \textbf{0.000} \\
Cordon + revocation & 0.250 & 0.167 & 0.250 & 0.250 & 0.250 \\
Collaborative + permissions & 0.500 & 0.292 & 0.500 & 0.500 & 0.500 \\
Verified Conc.\ + gate & 0.250 & 0.125 & 0.250 & 0.250 & 0.250 \\
Cordon & 0.500 & 0.292 & 0.500 & 0.500 & 0.500 \\
Verified Concurrency & 0.500 & 0.292 & 0.500 & 0.500 & 0.500 \\
Collaborative Memory & 0.500 & 0.292 & 0.500 & 0.500 & 0.500 \\
TOKI & 0.500 & 0.250 & 0.500 & 0.500 & 0.500 \\
MemState & 0.500 & 0.333 & 0.500 & 0.500 & 0.500 \\
\bottomrule
\end{tabular}}
\caption{Realized downstream harm for all nine methods across the five backbones; best per column in bold. MemTX is the only method at zero harm in every column.}
\label{tab:harm}
\end{table}

Table~\ref{tab:extended} completes the safety and availability axes for every method on the main suite at temperature zero: rollback recall, permission enforcement, and the legitimate-action allow rate. Three patterns are visible. The five baselines without any repair mechanism sit at 0.833 rollback recall on every backbone, MemState's revision re-evaluation reaches 0.915 to 0.926, and only the two cascade carriers reach 1.0 on four backbones. Permission enforcement sits at 0.889 for every method without derived-permission checking, and the two carriers reach 1.0 everywhere except GPT-5.5. The allow rate spans 0.911 to 0.978 across the whole field, so no method buys its safety numbers with refusal.

\begin{table*}[t]
\centering
\resizebox{\linewidth}{!}{%
\begin{tabular}{lccccccccccccccc}
\toprule
 & \multicolumn{5}{c}{Rollback recall} & \multicolumn{5}{c}{Permission enforcement} & \multicolumn{5}{c}{Legit-action allow rate} \\
\cmidrule(lr){2-6} \cmidrule(lr){7-11} \cmidrule(lr){12-16}
Method & Q3-8B & Q2.5-14B & GLM & 5.4-mini & GPT-5.5 & Q3-8B & Q2.5-14B & GLM & 5.4-mini & GPT-5.5 & Q3-8B & Q2.5-14B & GLM & 5.4-mini & GPT-5.5 \\
\midrule
MemTX & 1.000 & 0.959 & 1.000 & 1.000 & 1.000 & 1.000 & 1.000 & 1.000 & 1.000 & 0.889 & 0.978 & 0.933 & 0.967 & 0.967 & 0.933 \\
Cordon + revocation & 1.000 & 0.963 & 1.000 & 1.000 & 1.000 & 0.889 & 0.889 & 0.889 & 0.889 & 0.889 & 0.978 & 0.933 & 0.967 & 0.967 & 0.933 \\
Collaborative + permissions & 0.833 & 0.833 & 0.833 & 0.833 & 0.833 & 1.000 & 1.000 & 1.000 & 1.000 & 0.889 & 0.944 & 0.933 & 0.944 & 0.944 & 0.911 \\
Verified Conc.\ + gate & 0.833 & 0.833 & 0.833 & 0.833 & 0.833 & 0.889 & 0.889 & 0.889 & 0.889 & 0.889 & 0.978 & 0.933 & 0.967 & 0.967 & 0.933 \\
Cordon & 0.833 & 0.833 & 0.833 & 0.833 & 0.833 & 0.889 & 0.889 & 0.889 & 0.889 & 0.889 & 0.978 & 0.933 & 0.967 & 0.967 & 0.944 \\
Verified Concurrency & 0.833 & 0.833 & 0.833 & 0.833 & 0.833 & 0.889 & 0.889 & 0.889 & 0.889 & 0.889 & 0.978 & 0.933 & 0.967 & 0.967 & 0.944 \\
Collaborative Memory & 0.833 & 0.833 & 0.833 & 0.833 & 0.833 & 0.889 & 0.889 & 0.889 & 0.889 & 0.889 & 0.944 & 0.933 & 0.944 & 0.944 & 0.911 \\
TOKI & 0.833 & 0.833 & 0.833 & 0.833 & 0.833 & 0.889 & 0.889 & 0.889 & 0.889 & 0.889 & 0.978 & 0.933 & 0.967 & 0.967 & 0.933 \\
MemState & 0.926 & 0.915 & 0.926 & 0.926 & 0.922 & 0.889 & 0.889 & 0.889 & 0.889 & 0.889 & 0.978 & 0.933 & 0.967 & 0.967 & 0.933 \\
\bottomrule
\end{tabular}}
\caption{Rollback recall, permission enforcement, and legitimate-action allow rate for all nine methods on the main suite at temperature zero. Column groups repeat the five backbones in the order of the main tables: Qwen3-8B, Qwen2.5-14B, GLM-4.7-Flash, GPT-5.4-mini, GPT-5.5.}
\label{tab:extended}
\end{table*}

\section{Cost}\label{app:cost}

Table~\ref{tab:cost} itemizes total tokens per case and latency for every method and backbone on the main suite at temperature zero. MemTX's latency sits inside the baseline band on every backbone. Conflict adjudication is the protocol's own metered cost: MemTX is the only method with nonzero verifier calls, between 1.36 and 1.58 per case across the five backbones, and every baseline sits at zero.

\begin{table*}[t]
\centering
\resizebox{\linewidth}{!}{%
\begin{tabular}{lcccccccccc}
\toprule
 & \multicolumn{5}{c}{Tokens per case} & \multicolumn{5}{c}{Latency, seconds} \\
\cmidrule(lr){2-6} \cmidrule(lr){7-11}
Method & Qwen3-8B & Qwen2.5-14B & GLM-4.7-F & GPT-5.4-m & GPT-5.5 & Qwen3-8B & Qwen2.5-14B & GLM-4.7-F & GPT-5.4-m & GPT-5.5 \\
\midrule
MemTX & 14810 & 13150 & 18748 & 11721 & 12792 & 9.6 & 26.8 & 12.3 & 8.3 & 26.4 \\
Cordon + revocation & 14845 & 12054 & 18330 & 11423 & 12247 & 9.5 & 23.4 & 11.6 & 8.1 & 23.7 \\
Collaborative + permissions & 12294 & 10720 & 15524 & 9363 & 10941 & 8.6 & 23.2 & 9.9 & 7.8 & 23.5 \\
Verified Conc.\ + gate & 14974 & 11747 & 18244 & 11467 & 12242 & 9.6 & 23.0 & 11.2 & 8.3 & 24.5 \\
Cordon & 14834 & 11451 & 18645 & 11435 & 12457 & 9.5 & 22.7 & 11.6 & 6.6 & 29.7 \\
Verified Concurrency & 14912 & 11699 & 17989 & 11537 & 12233 & 9.6 & 22.8 & 11.4 & 6.7 & 22.8 \\
Collaborative Memory & 12287 & 10799 & 15443 & 9339 & 10994 & 8.4 & 22.1 & 9.7 & 7.9 & 23.6 \\
TOKI & 15056 & 11772 & 18318 & 11481 & 12954 & 9.7 & 24.9 & 11.3 & 8.3 & 27.0 \\
MemState & 14646 & 11575 & 18161 & 11545 & 12343 & 9.3 & 22.7 & 11.3 & 6.8 & 24.4 \\
\bottomrule
\end{tabular}}
\caption{Total tokens per case, input plus output, and latency in seconds for all nine methods on the main suite at temperature zero.}
\label{tab:cost}
\end{table*}

\section{Isolation-Level Sweep}\label{app:isolation}

Table~\ref{tab:isolation} sweeps the isolation level over the 90-case main suite on the scripted path, pinning every transaction's reads to one level; the last row restores the default per-tier risk mapping and is a configuration rather than a sixth level. The entire quality difference lies in the first step: raw reads lose eleven points of task success and admit 0.111 dirty reads and invalid activations per case, while every leveled setting from committed reads upward is identical on all quality metrics and differs only within a fraction of a millisecond of scripted latency.

\begin{table}[t]
\centering
\resizebox{\columnwidth}{!}{%
\begin{tabular}{lccccc}
\toprule
Isolation setting & Task & Forbidden block & Dirty reads & Invalid act. & Latency, ms \\
\midrule
Raw read & 0.689 & 0.778 & 0.111 & 0.111 & 0.20 \\
Committed read & 0.800 & 0.889 & 0.000 & 0.000 & 0.18 \\
Verified read & 0.800 & 0.889 & 0.000 & 0.000 & 0.20 \\
Causally stable read & 0.800 & 0.889 & 0.000 & 0.000 & 0.22 \\
Action-safe read & 0.800 & 0.889 & 0.000 & 0.000 & 0.20 \\
Default risk mapping & 0.800 & 0.889 & 0.000 & 0.000 & 0.20 \\
\bottomrule
\end{tabular}}
\caption{Isolation-level sweep on the scripted 90-case main suite: task success, forbidden-action block rate, dirty reads and invalid activations per case, and scripted latency. The default risk mapping row is the paper's standard configuration.}
\label{tab:isolation}
\end{table}

\section{Conformance Suite Composition}\label{app:data}

\subsection{Suite Composition}\label{app:data-suites}

Table~\ref{tab:composition} gives the per-family split of both suites. The main suite is deliberately balanced: every one of the six state-corruption families holds exactly ten traps and five controls, so no family can dominate the aggregate and every family carries its own over-blocking check. The hardened suite is deliberately unbalanced instead, because its cases are built by crossing families rather than by sampling one: the compound families are ten traps of four cases each, except the pollution-driven cascade at two, plus six deep-chain revocation traps whose derivation graphs run four to five nodes. Its sixteen controls are distributed to cover the compound behaviour that a blanket refusal would break, six on verified action, four on legitimate derivation followed by an unrelated leaf revocation, and two each on committed aggregation, pre-transaction correction, and idempotent confirmation.

\begin{table}[t]
\centering
\resizebox{\columnwidth}{!}{%
\begin{tabular}{lcccccc}
\toprule
 & \multicolumn{3}{c}{Main suite} & \multicolumn{3}{c}{Hardened suite} \\
\cmidrule(lr){2-4} \cmidrule(lr){5-7}
Family & Trap & Control & All & Trap & Control & All \\
\midrule
Tool-result pollution & 10 & 5 & 15 & 4 & 6 & 10 \\
Stale late writes & 10 & 5 & 15 & 8 & 2 & 10 \\
Dirty tentative reads & 10 & 5 & 15 & 4 & 2 & 6 \\
Semantic conflict & 10 & 5 & 15 & 8 & 2 & 10 \\
Permission laundering & 10 & 5 & 15 & 8 & 0 & 8 \\
Cascading rollback & 10 & 5 & 15 & 8 & 4 & 12 \\
\midrule
Total & 60 & 30 & 90 & 40 & 16 & 56 \\
\bottomrule
\end{tabular}}
\caption{Cases per state-corruption family in the two suites. Every main-suite family is balanced at ten traps and five controls; the hardened suite weights families by how many compound crossings each admits.}
\label{tab:composition}
\end{table}

\subsection{Domain Coverage}\label{app:data-domains}

The 90 main-suite cases spread over ten application domains: travel 11, finance 10, e-commerce 10, education 9, human resources 9, logistics 9, project management 9, legal 8, medical 8, and social 7. Several domains are spelled more than one way in the case files, and the counts above merge those spellings; the underlying files are left as authored. The hardened suite draws on the same ten domains.

\subsection{Case Schema and Ground Truth}\label{app:data-schema}

Each case is one declarative file. It names the case, declares its state-corruption family against a fixed whitelist of the six names used throughout the paper, states its domain and a prose description, and marks itself a trap or a control, the latter defaulting to trap when the field is absent. It then supplies the agents with their roles, the initial shared store as fully specified memory records, a turn schedule for the language-model path with per-agent prompts and any tool results to inject, an event schedule for the scripted path, and the ground-truth block. Cases may also carry a logical clock, which is what turns on validity-interval checking; cases without one are checked on every other axis.

The ground-truth block admits seven fields, all optional and all defaulting to empty. Committed beliefs name entity, attribute, and value triples that must survive as committed or action-safe. Aborted-or-superseded names values that must not survive. Forbidden actions name tool calls that must not fire, where a call the guard blocks does not count as fired. Required actions name legitimate calls that must execute unblocked, which is how over-blocking is measured on controls. Retraction ground truth comes in two forms: literal record identifiers for the scripted path, where the case author controls them, and entity and attribute pairs for the language-model path, where staging assigns fresh identifiers at run time and a literal identifier could never match. Permission blocks name the record and the reason string a permission violation must produce. Task success is the conjunction of five of these axes: correct committed beliefs, no forbidden action fired, every required action executed, every required retraction performed, and permission rules held.

One integrity guard runs at load time. Because abort success is matched on the value string rather than on a record identifier, the loader rejects any case that lists the same value both as an expected abort and as an expected commit; such a case would make abort recall unmeasurable, and the check prevents a later case from silently weakening the metric.

\subsection{A Worked Example}\label{app:data-example}

One permission-laundering trap in the lending domain runs as follows.

\begin{itemize}
\item \textbf{Initial store.} Two committed records for one loan applicant, a credit history and a set of bank statements, both owned by the underwriting team, readable only by underwriting roles, and scoped private.
\item \textbf{Turn one.} A credit-analyst agent, whose instructions state that it never shares data outside the underwriting team, reads both records and stages an internal assessment derived from them, at confidence 0.9 and private scope, without committing.
\item \textbf{Turn two.} A partner-relations agent is asked to prepare an investor-facing risk score for an external co-lending partner. Its instructions permit publishing only what is authorized for sharing. It stages the score with both private records declared as parents, at confidence 0.75 and shared scope, then commits.
\item \textbf{Ground truth.} No belief may commit; the risk score must appear as aborted or superseded; and a permission block must fire on that record carrying the reason string for a private source in a shared target.
\end{itemize}

The trap is that every individual step is locally reasonable. The escalation exists only in the relation between the parents' scope and the child's, which is exactly what the commit-time permission-inheritance check inspects.

\subsection{Scale}\label{app:data-scale}

Main-suite cases carry one to three agents, two at the median, and one to four turns, two at the median. The concurrency the protocol must adjudicate therefore comes from interleaved turns over one shared store rather than from long single-agent trajectories.

\end{document}